\pdfoutput=1
\documentclass[11pt]{article}

\usepackage{emnlp2021}

\usepackage{times}
\usepackage{latexsym}

\usepackage[T1]{fontenc}
\usepackage[utf8]{inputenc}

\usepackage{microtype}

\usepackage{booktabs}
\usepackage{graphicx}
\usepackage{enumitem}
\usepackage{amsmath}
\usepackage{color, colortbl}

\setlist{nosep} % remove space around itemize items

\setlist[itemize]{leftmargin=*} 
\usepackage{verbatim}  % comment out large chunks of text

\title{Text Detoxification using Large Pre-trained Neural Models }

\author{\textbf{David Dale}$^\ddag$, \textbf{Anton Voronov}$^{\ddag,\dag}$, \textbf{Daryna Dementieva}$^\ddag$, \textbf{Varvara Logacheva}$^\ddag$, \\ \textbf{Olga Kozlova}$^\dag$, \textbf{Nikita Semenov}$^\dag$, \textbf{and} \textbf{Alexander Panchenko}$^\ddag$ \\
$^\ddag$Skolkovo Institute of Science and Technology, Moscow, Russia\\
$^\dag$Mobile TeleSystems (MTS), Moscow, Russia\\
\href{mailto:a.panchenko@skoltech.ru}{\{d.dale,anton.voronov,daryna.dementieva,v.logacheva,a.panchenko\}@skoltech.ru} \\\href{mailto:a.panchenko@skoltech.ru}{\{oskozlo9,nikita.semenov\}@mts.ru}}

\date{}

\begin{document}
\maketitle
\begin{abstract}
We present two novel unsupervised methods for eliminating toxicity in text. Our first method combines two recent ideas: (1) guidance of the generation process with small style-conditional language models and (2) use of paraphrasing models to perform style transfer. We use a well-performing paraphraser guided by style-trained language models to keep the text content and remove toxicity. Our second method uses BERT to replace toxic words with their non-offensive synonyms. We make the method more flexible by enabling BERT to replace mask tokens with a variable number of words. 
Finally, we present the first large-scale comparative study of style transfer models on the task of toxicity removal. We compare our models with a number of methods for style transfer. The models are evaluated in a reference-free way using a combination of unsupervised style transfer metrics. Both methods we suggest yield new SOTA results. 
\end{abstract}

\section{Introduction}

Identification of toxicity in user texts is an active area of research \cite{zampieri-etal-2020-semeval,dsa-etal-2020-towards,han-tsvetkov-2020-fortifying}. The task of automatic rewriting of offensive content attracted less attention, yet it may find various useful applications such as making online world a better place by suggesting to a user posting a more neutral version of an emotional comment. The existing works on text detoxification~\cite{santos2018fighting,tran2020towards,civil_rephrases} cast this task as style transfer.  
The style transfer task is generally understood as rewriting of text with the same content and with altering of one or several attributes which constitute the ``style'', such as authorship~\cite{VOIGT18.903}, sentiment~\cite{Shen2017}, or degree of politeness~\cite{madaan-etal-2020-politeness}. %By ``content'', some authors denote the meaning of the sentence, while others interpret content preservation as keeping most of the words intact. 
Despite the goal of preserving the content, in many cases changing the style attributes changes the meaning of a sentence significantly.\footnote{For example, \newcite{lample-etal-2020-multiple} provide the following sentence as an example of transfer from male to female writing: \textit{Gotta say that beard makes you look like a Viking} $\rightarrow$ \textit{Gotta say that hair makes you look like a Mermaid.}} % (see Table~\ref{table:style_transfer_examples}). 
So in fact the goal of many style transfer models is to transform a sentence into a somewhat similar sentence of a different style on the same topic.\footnote{A formal  task definition is presented in Appendix~\ref{sec:definition}.}
We suggest that detoxification needs better preservation of the original meaning than many other style transfer tasks, such as sentiment transfer, so it should be performed differently.

We present two models for text detoxification, which have extra control for content preservation. The first model, \textbf{ParaGeDi}, is capable of fully regenerating the input. It is based on two ideas: external control of an output of a generation model by a class-conditioned LM~\cite{gedi} and formulation of style transfer task as paraphrasing~\cite{krishna-etal-2020-reformulating}. Being based on a paraphraser model, \textbf{ParaGeDi} explicitly aims at preserving the meaning of the original sentence. 
The second approach, \textbf{CondBERT}, inspired by \newcite{condbert}, follows the pointwise editing setup. It uses BERT to replace toxic spans found in the sentence with their non-toxic alternatives. The semantic similarity is maintained by showing the original text to BERT and reranking its hypotheses based on the similarity between the original words and their substitutes. Interestingly, BERT does not need any class-conditional pre-training to successfully change the text style from toxic to normal.

In addition, we perform a large-scale evaluation of style transfer models on detoxification task, comparing our new models with baselines and state-of-the-art approaches. We release our code and data.\footnote{\url{https://github.com/skoltech-nlp/detox}}

Our contributions are as follows:

\begin{itemize}[noitemsep]
    \item We propose two novel detoxification methods based on pre-trained neural language models: \textbf{ParaGeDi} (paraphrasing GeDi) and \textbf{CondBERT} (conditional BERT).
    
    \item We conduct an evaluation of these models and their comparison with a number of state-of-the-art models for text detoxification and sentiment transfer and release the detoxification dataset.
    
    \item We create an English parallel corpus for the detoxification task by retrieving toxic/safe sentence pairs from the ParaNMT dataset~\cite{wieting2017paranmt}. We show that it can further improve our best-performing models.
\end{itemize}

\section{Related Work}
\label{section:related}

One of the most straightforward ways of solving style transfer task is to ``translate'' a source sentence into the target style using a supervised encoder-decoder model~\cite{gyafc}. Since the source and the target are in the same language, pre-trained LMs such as GPT-2~\cite{gpt2} can be applied for this task --- fine-tuning them on relatively small parallel corpora gives a good result~\cite{harnessing}. However, this method is used quite rarely because of the lack of sufficiently large parallel data. The rest of described models are trained in an unsupervised way. 

\paragraph{Pointwise Editing Models}

A relatively easy yet efficient style transfer method is to leave the sentence intact and manipulate only individual words associated with the style. Delete-Retrieve-Generate (DRG) framework \cite{DRG} was the first effort to perform such transfer. It proposes four methods based on this principle. Two of them perform well on our data: \textbf{DRG-RetrieveOnly} retrieves a sentence with the opposite style which is similar to the original sentence and returns it, and \textbf{DRG-TemplateBased} takes the style attributes from it and plugs them into the original sentence.
Here, the performance depends on the methods for the identification of style markers and retrieval of replacements. Words associated with style are typically identified either based on their frequencies as in the original paper, some works use attention weights as features \cite{sudhakar2019transforming}.

Alternatively, style transfer can use Masked Language Modelling (MLM). An MLM trained on a dataset with style labels picks a %most suitable 
replacement word based not only on the context, but also on the style label. An example of such model is \textbf{Mask \& Infill}~\cite{MLM1}. It is most similar to \textbf{CondBERT} method we propose. However, CondBERT performs additional control over the style and the content preservation and is able to make multi-word replacements.
Another similar model of this type is described by \newcite{MLM2}. It has a more complicated structure: there, two MLMs trained on corpora of different styles perform replacements jointly.

\paragraph{End-to-end Architectures}

In contrast to these models, there exist end-to-end architectures for style transfer. They encode the source sentence, then manipulate the resulting hidden representation in order to incorporate a new style, and then decode it. Some of them disentangle the hidden representation into the representation of content and style~\cite{disentangled}. The others force the encoder to represent style-independent content~\cite{toward}.
Alternatively, the model \textbf{DualRL} by \newcite{luo2019dual} performs a direct transfer from the source to the target style. The task is paired with the dual task (back transfer to the source style) which allows models to train without parallel data. The Deep Latent Sequence Model (\textbf{DLSM}) model by \newcite{he2020a} uses amortized variational inference to jointly train models for the primal and dual tasks. The Stable Style Transformer (\textbf{SST}) method~\cite{lee2020stable} trains a pair of sequence-to-sequence transformers for primal and dual tasks using cross-entropy of a pretrained style classifier as an additional discriminative loss. 
The Style Transfer as Paraphrase (\textbf{STRAP}) method by~\newcite{krishna-etal-2020-reformulating} views a style transfer model as a paraphraser that adds attributes of a particular style to a text. The authors create pseudo-parallel data by transferring style-marked texts to neutral with a pre-trained general-purpose paraphraser and then train sequence-to-sequence models on these neutral-to-styled parallel datasets.
Our \textbf{ParaGeDi} model is conceptually similar to these methods. However, unlike these methods, the style is not infused into the model or a sentence representation but is imposed on the generator by another model.

\paragraph{Detoxification}

Detoxification of text is a relatively new style transfer task. The first work on this topic by \cite{santos2018fighting} is an end-to-end seq2seq model trained on a non-parallel corpus with autoencoder loss, style classification loss and cycle-consistency loss.
A more recent work by \newcite{tran2020towards} uses a pipeline of models: a search engine finds non-toxic sentences similar to the given toxic ones, an MLM fills the gaps that were not matched in the found sentences, and a seq2seq model edits the generated sentence to make it more fluent. Finally, \newcite{civil_rephrases} detoxify sentences by fine-tuning T5 as a denoising autoencoder with additional cycle-consistency loss.
\newcite{pplm} and \newcite{gedi} approach a similar problem: preventing a language model from generating toxic text. They do not need to preserve the meaning of the input text. 
However, the idea of applying a discriminator to control an LM during generation can be used for style transfer, as we show in our experiments.

\section{Paraphrasing GeDi Model}

The recently proposed GeDi model~\cite{gedi} performs text generation from scratch guided by a language model informed about specific attributes of a text, e.g. style or topic. We extend this model by enabling it to paraphrase the input text. 
\subsection{GeDi}

The original GeDi model consists of two components: a generation model (GPT-2) and a discrimination model, which is also a GPT-2 trained on sentences with additional sentence-level style labeling --- during training the style label is prepended to a sentence. This makes the discriminating model learn the word distributions conditioned on a particular label. 
At each generation step, the distribution of the next token predicted by the main model $P_{LM}$ is modified using an additional class-conditional language model $P_D$ and the Bayes rule:

\vspace{-7mm}
 $$   P(x_t|x_{<t}, c) \propto P_{LM}(x_t|x_{<t}) P_D(c|x_t, x_{<t}) $$
\vspace{-7mm}

Here, $x_t$ is the current token, $x_{<t}$ is the prefix of the text, and $c$ is the desired attribute (e.g. toxicity or sentiment) --- one of $C$ classes. The first term is produced by the main language model $P_{LM}$, and the second term is calculated using the Bayes rule and the additional class-conditional language model $P_{CC}$. Thus, the tokens which are more likely to appear in a text of the chosen style get a higher probability:

\vspace{-7mm}
$$ P_D(c|x_t, x_{<t}) \propto P(c) P_{CC}(x, x_{<t}|c) $$
\vspace{-7mm}

The name GeDi stands for Generative Discriminator, because a language model, which is generative by its nature, is used as a discriminator for guiding the generation process.
GeDi was successfully applied to guiding a GPT-2 language model towards generating texts of particular topics and making the generated text less toxic.

\subsection{ParaGeDi} 
\label{section:ParaGeDi}

\begin{figure}
    \centering
    \includegraphics[width=1.0\linewidth]{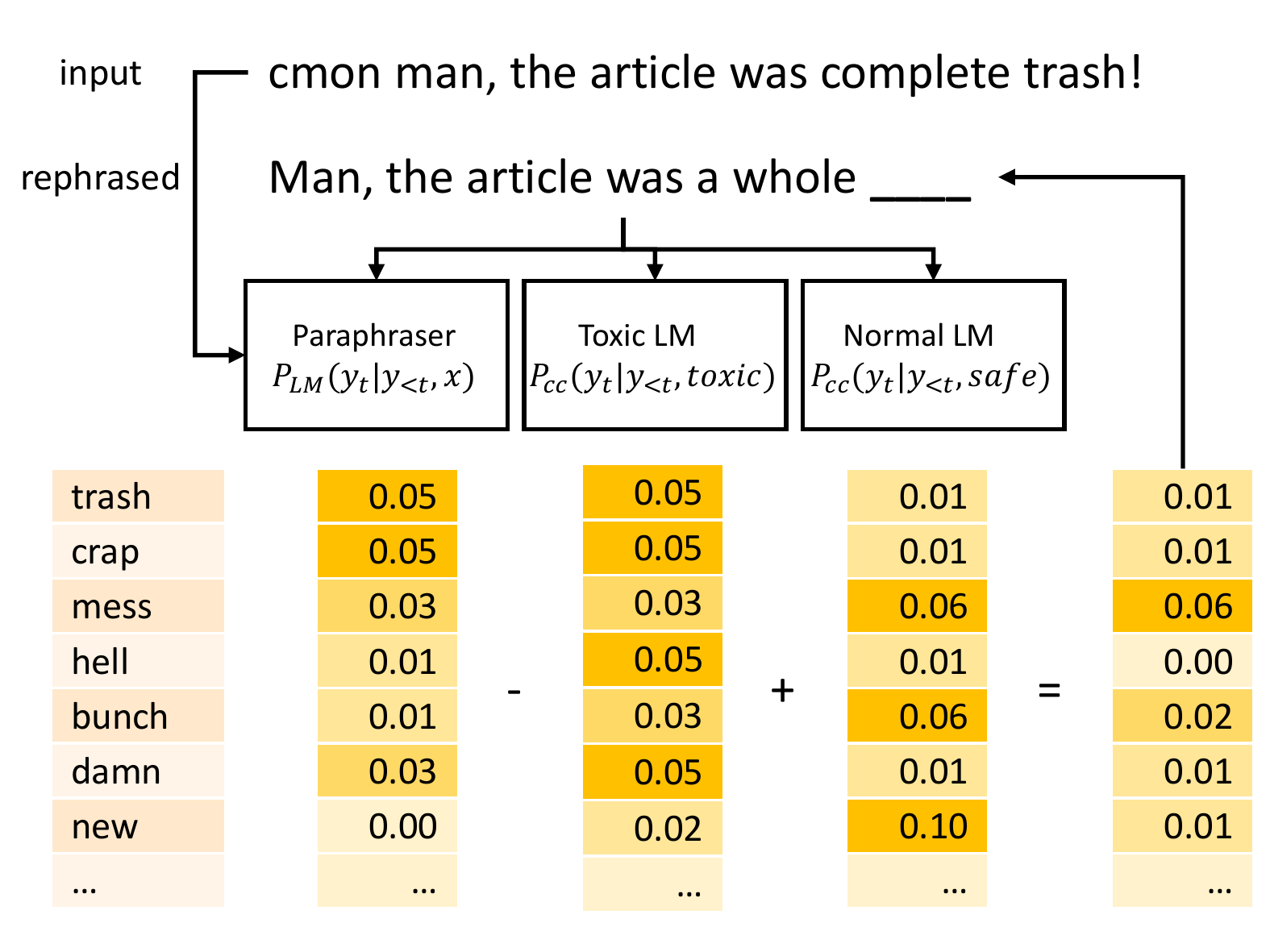}
    \caption{The overview of ParaGeDi model.}
    \label{fig:para_gedi_example}
\end{figure}

In order to enable GeDi to preserve the meaning of the input text, we replace the regular language model in it with a model capable of paraphrasing. 
If we denote the original text by $x$, the generated text of length $T$ by $y$, and the desired style by $c$, ParaGeDi models the following probability: 

\vspace{-5mm}

\begin{equation}
    \begin{split}
P(y_t|y_{<t}, x, c) \propto P_{LM}(y_t|y_{<t},x) P(c|y_t, y_{<t},x) \\       \approx P_{LM}(y_t|y_{<t},x) P_{D}(c|y_t, y_{<t})
    \end{split}
    \notag
\end{equation}
\vspace{-8mm}

The last step is an approximation because the class probability should be conditioned on both $x$ and $y$. %However, we decide to condition the class only on the output text by analogy with the original model. 
%, because the text $x$, being a paraphrase of $y$, contains the information about the suffix of $y$ not yet generated, and therefore about its class $c$. 
However, this approximation, although not being fully justified, allows us to decouple the paraphraser model (which requires a parallel corpus for training) from the style model (which requires only texts with style labels, not necessarily parallel). %The two models only need to share a common vocabulary. 
The paraphraser and the style model can be trained independently. Moreover, we can plug in any paraphraser as long as it shares the vocabulary with the class-conditional LM.
%Because paraphrasing seems to be more difficult than controlling style, we suggest using an existing good pretrained paraphraser, and train the style-conditional LM using the vocabulary of the paraphraser.
The third (optional) component of this model is a reranker --- an external model which reweighs the hypotheses generated by the style LM-guided paraphraser with respect to the style. Our reranker is a pre-trained toxicity classifier which chooses the least toxic hypothesis generated by the ParaGeDi model. Figure~\ref{fig:para_gedi_example} illustrates the workflow of our model.

ParaGeDi is trained as follows. Its loss $\mathcal{L}_{ParaGeDi}$ consists of a linear combination of two losses: the generative loss $\mathcal{L}_{G}$ used in LM training, and the discriminative loss $\mathcal{L}_{D}$ which further pushes different classes away from one another.

\vspace{-5mm}

$$\mathcal{L}_{G}=-\frac{1}{N} \sum_{i=1}^{N}\frac{1}{{T}_{i}}\sum _{t=1}^{T_i}\log P(y_t^{(i)}|y_{<t}^{(i)}, c^{(i)}) $$
 
 \vspace{-2mm}

$$\mathcal{L}_{D}=-\frac{1}{N} \sum_{i=1}^{N}\log P(c^{(i)}|y^{(i)}_{1:T_i}) $$

\vspace{-2mm}

$$\mathcal{L}_{ParaGeDi} = \lambda \mathcal{L}_{D} + (1-\lambda) \mathcal{L}_{G} $$

We enhance the model with a number of inference heuristics that improve content preservation and increase the style transfer accuracy. First, we use a heuristic from the original GeDi model. We raise the conditional LM probability to the power $w>1$, which biases the discriminator towards the correct class during generation: 
\begin{equation}
P(y_t|y_{<t}, x, c) \propto P_{LM}(y_t|y_{<t},x) P_{CC}(c|y_t, y_{<t})^w
\notag
\end{equation}

Besides that, we suggest two new heuristics:

\textbf{Smoothing of probabilities} --- adding a small $\alpha>0$ to all probabilities from the conditional language model discourages the generation of tokens with low probability conditional on all classes:
\begin{equation}
            P_{\alpha}(c|x_t, x_{<t}) 
            = \frac{\alpha + P(c) P_{CC}(x, x_{<t}|c)}{\sum_{c^{'}\in C}\left( \alpha + P(c^{'}) P_{CC}(x, x_{<t}|c^{'})\right)}
\notag
\end{equation}        

\textbf{Asymmetric lower and upper bounds} ($l$ and $u$) for class-conditional corrections:
\begin{equation} 
P_{\alpha,l,u}(c|x_t, x_{<t}) = \max(l, \min(u, P_{\alpha}(c|x_t, x_{<t}))).
\notag
\end{equation}

By decreasing the value of $u$ we discourage the insertion of new tokens, as opposed to prohibiting existing tokens. For the problem of detoxification, it means that the model will try less to insert polite words than to delete toxic words from the sentence.

\section{Conditional BERT Model}

BERT \cite{devlin-etal-2019-bert} has been trained on the task of filling in gaps (``masked LM''), we can use it to insert non-toxic words instead of the toxic ones. This approach has been suggested by \newcite{condbert} as a method of data augmentation. 
The authors identify words belonging to the source style, replace them with the \texttt{[MASK]} token, and the BERT model then inserts new words of the desired style in the designated places. To push BERT towards the needed style, the authors fine-tune BERT on a style-labelled dataset by replacing segmentation embeddings of original BERT with trainable style embeddings.

\begin{figure}[ht!]
    \centering
   \includegraphics[width=1.0\linewidth]{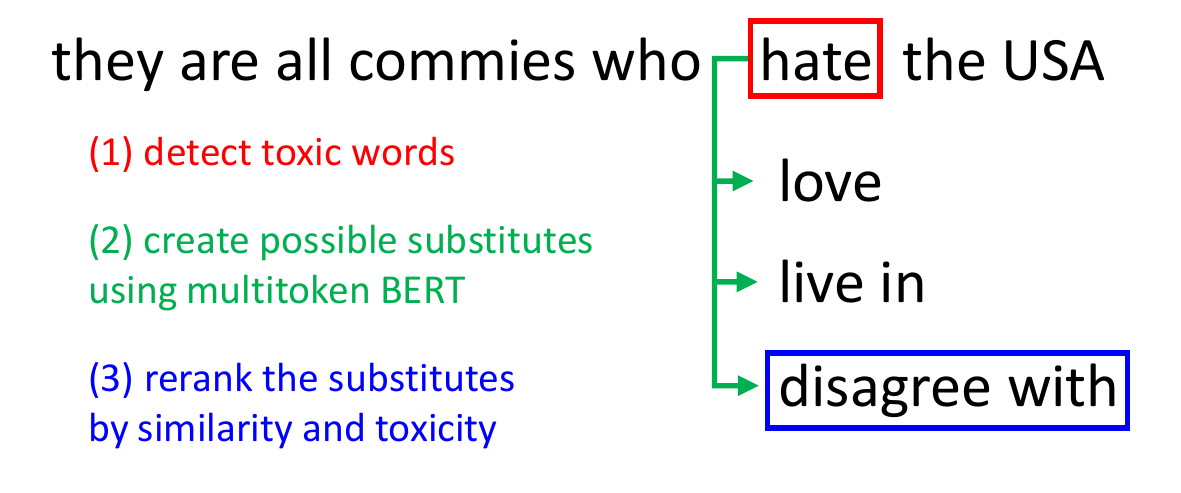}
    \caption{The overview of the CondBERT model.}
    \label{fig:para_condbert_example}
\end{figure}

We perform some changes to this model to adapt it for the detoxification task. While in the original conditional BERT model the words are masked randomly, we select the words associated with toxicity. This can be done in different ways, e.g. by training a word-level toxicity classifier or manually creating a vocabulary of rude and toxic words. We use a method which does not require any additional data or human effort. We train a logistic bag-of-words toxicity classifier. This is a logistic regression model which classifies sentences as toxic or neutral and uses their words as features. As a byproduct of the training process, each feature (word) yields a weight which roughly corresponds to its importance for classification. The words with the highest weights are usually toxic. We use the normalised weights from the classifier as toxicity score.
%A method that does not require annotations of toxic spans is a toxicity classifier. We train sentence-level logistic toxicity classifier. We train such a classifier
%and associate toxicity of a word with its weight in the classifier. 
The overview of CondBERT is shown in Figure \ref{fig:para_condbert_example}.

For each word in a sentence, we compute the toxicity score and then define toxic words as the words with the score above a threshold $t=max(t_{min}, max(s_1, s_2, ..., s_n)/2)$, where $s_1, s_2, ..., s_n$ are scores of all words in a sentence and $t_{min}=0.2$ is a minimum toxicity score. This adaptive threshold allows balancing the percentage of toxic words in a sentence so that we avoid cases when too many or no words are marked as toxic.% \footnote{Our model does not perform toxicity detection. Therefore, we assume that all input sentences are toxic and have to contain toxic words, and low toxicity scores should not be considered a sign of the absence of toxicity.} 

To preserve the meaning of the replaced word, we employ the content preservation heuristics suggested 
by \newcite{arefyev2020always}: (i) Preserve the original tokens instead of masking them before the replacement; (ii) Rerank the replacement words suggested by BERT by the similarity of their embedding with the embedding of the original word.

Despite using class-specific sentence embeddings, conditional BERT often predicts toxic words, apparently paying more attention to the context than to the embeddings of the desired class. To force the model to generate non-toxic words we calculate the toxicity of each token in BERT vocabulary and penalize the predicted probabilities of tokens with positive toxicities.

Finally, we enable BERT to replace a single \texttt{[MASK]} token with multiple tokens. We generate each next token progressively by beam search and score each multitoken sequence by the harmonic mean of the probabilities of its tokens. 

\section{Detoxification Experiments}

We train the two new models and a number of other systems for text detoxification. Below we describe datasets, evaluation setup, and results.

\subsection{Toxicity Classifier}
\label{section:clf}

We train two binary classifiers of toxicity. One of them is used to rerank hypotheses in the ParaGeDi model, and the other participates in the evaluation. We train these two classifiers on different sets of data. The overall dataset is the merge of the English parts of the three datasets by Jigsaw~\cite{jigsaw_toxic,jigsaw_bias,jigsaw_multi}, containing around 2 million examples. We split it into two parts and fine-tune a RoBERTa model \cite{roberta} on it. We use the \texttt{roberta-large} model from the original repository. The classifiers perform closely on the test set of the first Jigsaw competition, reaching the AUC-ROC of 0.98 and F$_1$-score of 0.76.

\subsection{Dataset}

For training and testing of the style transfer models, we use the English data from the first Jigsaw competition~\cite{jigsaw_toxic}. The majority of our methods are trained on non-parallel corpora of source and target styles. To prepare the toxic dataset, we divide the comments labelled as toxic into sentences (the original comments are often too long) and classify each of them with our toxicity classifier. Sentences classified as toxic are used as the toxic part of the dataset (we find 154,771 of them). To select the neutral part of the dataset, we randomly pick the same number of non-toxic sentences from the sentence-separated Jigsaw data. The test set is prepared analogously to the test set of the Jigsaw competition: we use 10,000 sentences with the highest toxicity score according to our classifier.

\subsection{Metrics}
\label{section:metrics}

There is no parallel test set available for the detoxification task, so we cannot use BLEU, METEOR or ROUGE metrics and resort to referenceless evaluation. 
Style transfer models need to change the style, preserve content and produce a fluent text. These parameters are often inversely correlated, so we need a compound metric to find a balance between them. We follow the evaluation strategy of \newcite{krishna-etal-2020-reformulating} and use the metric \textbf{J}, which is the multiplication of sentence-level \textit{style accuracy}, \textit{content preservation}, and \textit{fluency}. The system-level \textbf{J} is the average of sentence-level scores. \textit{Style accuracy} (ACC) is measured with a pre-trained toxicity classifier described in Section \ref{section:clf}. \textit{Content preservation} (SIM) is evaluated as the similarity of sentence-level embeddings of the original and transformed texts computed by the model of \newcite{wieting2019beyond}. \textit{Fluency} (FL) measured with the classifier of linguistic acceptability trained on the CoLA dataset~\cite{cola}. \textbf{J} is computed as the average of their sentence-level product. In addition to that, we tried a similar aggregated metric \textbf{GM}~\cite{pang2018unsupervised,civil_rephrases} which uses perplexity as the measure of fluency and employs a different aggregation method. Our preliminary experiments showed that \textbf{J} and \textbf{GM} are strongly correlated, so we keep only \textbf{J} for further evaluation.

\subsection{Implementation Details}
\label{section:implementation}
For ParaGeDi, we use a pre-trained T5-based \cite{T5} paraphraser,\footnote{\url{https://huggingface.co/ceshine/t5-paraphrase-paws-msrp-opinosis}} fine-tuned on a random subsample of the ParaNMT dataset \cite{wieting2017paranmt}. As a discriminator, we fine-tune the \verb|gpt2-medium| model \cite{gpt2} on the training part of the Jigsaw-1 dataset using two control codes for toxic and polite texts. Before fine-tuning, we change the vocabulary of the discriminator to match that of T5, and update its embeddings accordingly. We train the discriminator using a combined generative and discriminative loss from \newcite{gedi}, adapting their code for this purpose.

We use beam search decoding with 10 beams to generate paraphrase candidates with the paraphraser and discriminator described above. We apply the classifier from section \ref{section:clf} to select the least toxic candidate from the 10 resulting paraphrases.

\subsection{Competing Methods}

We compare our models with state-of-the-art methods described in Section~\ref{section:related}: DRG-TemplateBased, DRG-RetrieveOnly, Mask\&Infill, DLSM, STRAP, and SST. We also implement three other baselines: Machine Translation, Detoxifying GPT-2, and Paraphraser. We do not directly compare our models with GeDi, because it is a language model and was not explicitly trained to transform texts. 

\textbf{Machine Translation}~~There is evidence that automatic translation tends to eliminate toxicity~\cite{prabhumoye-etal-2018-style}. 
Thus, we use a chain of Machine Translation models for detoxification. Namely, we perform English~$\rightarrow$~Pivot~$\rightarrow$~English translation. 
We choose French and Igbo as pivot languages. French is resource-rich and structurally similar to English, which ensures high-quality translations. Conversely, Igbo is low-resourced and syntactically different. Both experiments are conducted using Google Translate API.

\begin{table*}[ht!]
\footnotesize
\centering
\begin{tabular}{l|l|l|l|l}
\toprule

 \bf Model       & \bf ACC & \bf  SIM &  \bf  FL & \bf  J \\
 \midrule
 \rowcolor{green!30}CondBERT (ours)  & 0.94 &	0.69	& 0.77 & 0.50 $\pm$ 0.0037* \\
 \rowcolor{green!30}ParaGeDi (ours) & 0.95 &	0.66 &	0.80 &	0.50 $\pm$ 0.0032* \\
 %\midrule
 Mask\&Infill \cite{MLM1} & 0.78 &	0.80 &	0.49 &	0.31 $\pm$ 0.0041 \\
 DRG-TemplateBased \cite{DRG} &0.66 &	0.82 &	0.59 &	0.30 $\pm$ 0.0041 \\
 DRG-RetrieveOnly \cite{DRG} &0.93&	0.33&	0.84& 0.26 $\pm$ 0.0019 \\
 DLSM \cite{he2020a} & 0.62  &	0.72  &	0.48 &	0.17 $\pm$ 0.0033 \\
 \rowcolor{gray!30} Detoxifying GPT-2 (baseline) & 0.54 &		0.48  &	0.72 &	0.17 $\pm$ 0.0026 \\
 STRAP \cite{krishna-etal-2020-reformulating} & 0.29  &	0.69  &	0.80	& 0.15 $\pm$ 0.0027 \\
 \rowcolor{gray!30} En$\rightarrow$Ig$\rightarrow$En MT (baseline) & 0.37	 &	0.68 	& 0.57 	& 0.12 $\pm$ 0.0025 \\
\rowcolor{gray!30}  T5 paraphraser (baseline) & 0.15  &	0.90 &		0.87 &	0.11 $\pm$ 0.0029 \\
 SST \cite{lee2020stable} &  0.80 &		0.55 &		0.12 &	0.05 $\pm$ 0.0019 \\
 \rowcolor{gray!30} En$\rightarrow$Fr$\rightarrow$En MT (baseline) & 0.06 &	0.91 &		0.81 &	0.04 $\pm$ 0.0019 \\ 
 %\rowcolor{gray!50} Original texts & 0.00 &	1.00  &	313.74 &	0.83 &	0.000 &	0.00 \\ 
 \bottomrule
\end{tabular}
\caption{Performance of detoxification models. Gray lines denote baselines, green lines denote models suggested in this work. The models are sorted with respect to the aggregated \textbf{J} score. The asterisk * denotes the scores that are significantly higher than the third best model (Mask\&Infill) with $p<0.01$, based on the paired \textit{t-test}.}
\label{tab:results}
\end{table*}

\textbf{Detoxifying GPT-2}~~GPT-2 \cite{gpt2} can be adapted to a wide range of NLP tasks using a very small task-specific dataset. We experiment with the model's ability to perform sequence-to-sequence tasks. We train it on a parallel dataset of 200 toxic and safe sentences. We randomly select toxic sentences from the Google Jigsaw toxic comments dataset \cite{jigsaw_toxic} and manually rewrite them in the neutral tone. % eliminating toxicity. 

\textbf{Paraphraser}~~\newcite{krishna-etal-2020-reformulating} suggest that a general-purpose paraphraser can remove style markers from text. We check this assumption.

\subsection{Results}
\label{section:results}

The performance of all tested models is given in Table \ref{tab:results}. Both \textbf{ParaGeDi} and \textbf{CondBERT} outperform other models by a large margin. The success of CondBERT is explained by its use of heuristics targeted at the components of the metric: (i)~it is penalized for generating toxic tokens, which ensures a high ACC score, (ii)~over 80\% tokens stay unchanged, and the replacements are selected with respect to the similarity to the original words, increasing the overall SIM score, (iii)~MLM is pre-trained to replace masked tokens with plausible substitutes, increasing FL. ParaGeDi is behind in terms of similarity but has a slightly higher fluency because generation is a better strategy in terms of text naturalness than pointwise corrections.
The closest competitor of our models is Mask\&Infill which uses similar principles as CondBERT. However, some engineering decisions (e.g. masking of all words at once) result in a substantial drop in fluency and some decrease in style transfer accuracy.

Surprisingly, many advanced models perform below the simplistic (DRG) models \textbf{TemplateBased} and \textbf{RetrieveOnly}. TemplateBased achieves a high similarity because it keeps most of the original sentence intact, and RetrieveOnly yields a high similarity and style transfer accuracy, because it outputs real non-toxic sentences from the training data.
\textbf{DLSM} and \textbf{SST} models perform full re-generation of text (as opposed to pointwise corrections). More importantly, their decoders are trained from scratch on a relatively small dataset, hence their low fluency scores. Conversely, \textbf{STRAP}, which also generates the sentence, has the access to the larger pseudo-parallel data, resulting in higher fluency.

Another finding is that \textbf{MT} has detoxification ability. However, it is inversely correlated with its quality: the En$\rightarrow$Ig$\rightarrow$En detoxifies 37\% of sentences but has low SIM and FL scores. Conversely, En$\rightarrow$Fr$\rightarrow$En yields a better output which keeps most of the original features, including toxicity. The same applies to the T5 \textbf{paraphraser}. %, which has very high SIM and FL scores and very low ACC. 
On the other hand, the \textbf{GPT-2} model can be trained to detoxify even on a very small number of parallel sentences (200 in our experiments). Although it performs below many other models, we suggest that training it on a larger parallel dataset can boost its performance. We show examples of the paraphrases by the best-performing models in Table~\ref{table:detox_examples}.

Additional examples and qualitative analysis can be found in Appendices~\ref{appendix:detox_examples} and~\ref{sec:qualitative}, respectively.

\begin{table*}[htb]
\centering
\footnotesize
\begin{tabular}{ll}
\toprule
Input & no room for those stupid antics . \\
ParaGeDi (ours) & There's no space for all those adversities. \\
CondBERT (ours) & no room for those . . . those bizarre exploits . \\ 
Mask\&Infill & no room for those compelling antics . \\
DRG-TemplateBased & no room for those conventions antics . \\
\midrule
Input & zoological conspiracy theories ( arabisraeli conflict ) , by contrast , sucks horribly . \\
ParaGeDi (ours) & By contrast, the zoological theories of conspiracy (the Arab-Israeli conflict) suffer horribly. \\
CondBERT (ours) & zoological conspiracy theories ( arabisraeli conflict ) , by contrast , worked out badly . \\ 
Mask\&Infill & zoological conspiracy theories ( arabisraeli conflict ) , by contrast , ends horribly . \\
DRG-TemplateBased & zoological conspiracy theories ( arabisraeli conflict ) , by contrast , instrumental horribly . \\
\midrule
Input & this whole article is bullshit . \\
ParaGeDi (ours) & This article's not a good deal. \\
CondBERT (ours) & this whole article is considered incomplete . \\ 
Mask\&Infill & this whole article is compelling .\\
DRG-TemplateBased & this whole article is substantially . \\
%input & yeh , so you like to hunt prostitutes for your own pleasure . \\
%ParaGeDi (ours) & Yeah, so you'd like to find prostitute-hunters for the sake of it. \\
%CondBERT (ours) & yeh , so you like to hunt other people just for your own personal enjoyment . \\ 
%Mask\&Infill & yeh , so you like to hunt fish for your own pleasure . \\
%DRG-TemplateBased & yeh , so you like to hunt arrangements for your own pleasure . \\
%\midrule
%input & trump hasnt done anything but demonstrate his arrogance and ignorance . \\
%ParaGeDi (ours) & Trump has done nothing, but showed his arrogance and mistrust. \\
%CondBERT (ours) & trump hasnt done anything but demonstrate his prejudices and his prejudice . \\
%Mask\&Infill & trump hasnt done anything but demonstrate his arrogance and compelling  . \\
%DRG-TemplateBased & trump hasnt done anything but demonstrate his arrogance and ignorance  symbolic . \\
%\midrule
% input & they are willing to kill john mccain so he can help them kill thousands more . \\
%ParaGeDi (ours) & They'd be willing to take John McCain so he can help them with thousands more. \\
%CondBERT (ours) & they are willing to give guns to john mccain so he can help them to save many thousands more . \\ 
%Mask\&Infill & they are willing to compelling john mccain so he can help them achieve  thousands more . \\
%DRG-TemplateBased & they are willing to kill john mccain so he can help them kill thousands more  disclosure . \\
\bottomrule
\end{tabular}
\caption{Examples of detoxification by different models.}
\label{table:detox_examples}
\end{table*}

\subsection{Parameter Selection }

Our models use multiple parameters and heuristics. We perform an ablation study to explore their usefulness. It turns out that the crucial features of CondBERT are multiword replacement which ensures high fluency and toxicity penalty which increases style strength. On the other hand, masking of all tokens at once as well as control of similarity do not affect the quality. More details on the CondBERT ablation study are given in Appendix~\ref{appendix:condbert}.

\begin{figure}[ht!]
    \centering
   \includegraphics[width=1.0\linewidth]{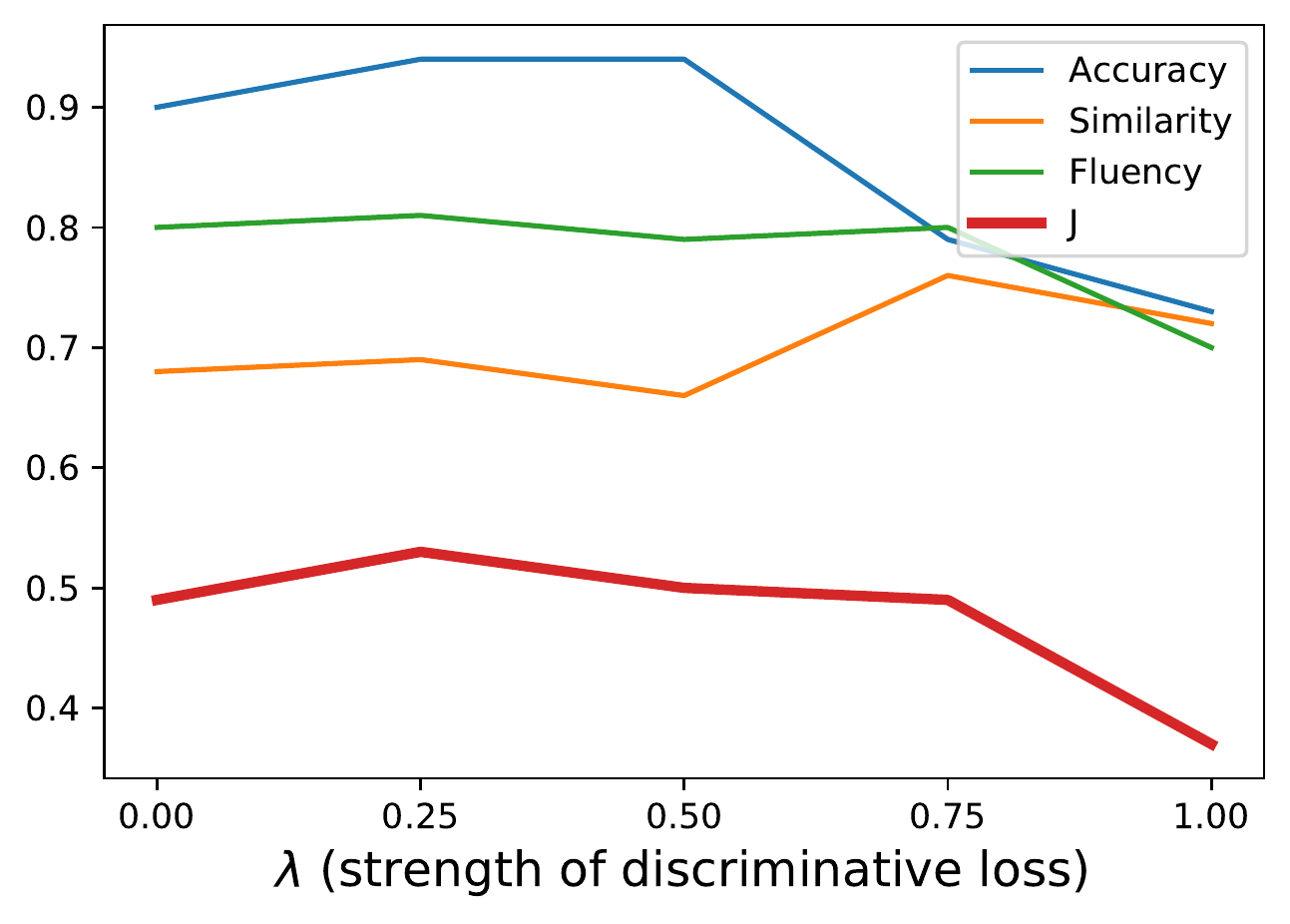}
    \caption{Performance of ParaGeDi with the varying $\lambda$ parameter (greater $\lambda$ corresponds to the stronger influence of the discriminative loss and smaller $\lambda$ means the stronger influence of the generative loss).}
    \label{fig:lambda}
\end{figure}

ParaGeDi has only one training hyperparameter $\lambda$ which controls the strength of its discriminative loss. We discover its value has only a marginal effect on the overall quality: the value of \textbf{J} decreases only for $\lambda=1$ which constitutes the absence of generative loss (see Figure \ref{fig:lambda}). The style strength control influences the style accuracy, whereas the use of word probability upper bound increases the similarity, and the absence of beam search decreases fluency. On the other hand, reranking, beam size, smoothing do not affect the model performance. An ablation study of the ParaGeDi model can be found in Appendix~\ref{appendix:paragedi}.

%\subsection{Sentiment transfer experiments}
%Text detoxification is not as well-established as other style transfer tasks, which makes it is difficult to put our models in the context of other works on style transfer. Thus, we conduct an experiment on a different domain, namely, sentiment transfer. In terms of the \textbf{J} metric, ParaGeDi outperforms all the baselines, primarily due to the outstanding fluency. The details are given in Appending \ref{app:sentiment}.

\section{Mining a Parallel Detoxifying Corpus}
\label{section:parallel}
The STRAP model~\cite{krishna-etal-2020-reformulating} is based on the assumption that a regular paraphraser can transform a stylistically marked text into a neutral text. Although our experiments show that a paraphraser, as well as an MT model, are bad detoxifiers on their own (see Section~\ref{section:results}), we suggest that it is possible to find occasional detoxifying sentence pairs in a large parallel dataset of paraphrases. 

\textbf{Experimental Setup}~~To test this hypothesis, we classify the sentences from the ParaNMT paraphrase dataset \cite{wieting2017paranmt} with our toxicity classifier described in Section~\ref{section:clf} and obtain 500,000 paraphrase pairs where one sentence is more toxic than the other (for more details on the data collection process please see Appendix~\ref{appendix:parallel}). %We publish this dataset.  
We then compare the regular paraphraser from Section \ref{section:implementation} fine-tuned on a random subset of ParaNMT (\textbf{regular}) with its version fine-tuned on the mined toxic/safe parallel paraphrase corpus (\textbf{mined}). We also plug both paraphrasers into ParaGeDi model and compare the overall performance. The results are shown in Table~\ref{tab:parallel}. 

\begin{table}
\centering
\small

\begin{tabular}{l|l|l|l|l}
\toprule
\bf Model     & \bf ACC      & \bf SIM     &\bf  FL     & \bf  J      \\ \midrule
\multicolumn{5}{c}{Paraphraser} \\ \midrule
regular &      0.15 & 0.90 & 0.87 & 0.11 $\pm$0.003      \\
mined &      0.42 & 0.87 & 0.91 & 0.31 $\pm$0.004      \\ \midrule
\multicolumn{5}{c}{ParaGeDi}     \\ \midrule
regular &      0.94 & 0.66 & 0.77 & 0.50 $\pm$0.003      \\
mined     &      0.98 & 0.66 & 0.84 & 0.54 $\pm$0.003      \\ \bottomrule
    
\end{tabular}
\caption{Comparison of paraphrasers for ParaGeDi.}
\label{tab:parallel}

\end{table}

\textbf{Discussion of Results}~~None of the paraphrasers can fully detoxify the test set, but the \textbf{mined} paraphraser gets a better ACC than the \textbf{regular} one (42\% vs 15\%). When we replace the regular paraphraser with the detoxifying one in \textbf{ParaGeDi}, both detoxification rate and fluency improve without loss in the similarity score. This leaves us with the \textbf{J} score of 0.54, which is the highest score we obtained in our detoxification experiment. We do not include it in the main results (Table~\ref{tab:results}) because this model is not unsupervised. However, this result shows that the general-purpose ParaNMT corpus contains a large number of toxic/safe paraphrase pairs. We believe that mining parallel training sets from large corpora, as opposed to unsupervised methods of style transfer, is a fruitful direction.

\section{Human Evaluation of Detoxification}

%It has been shown that the automatic metrics 
While the automatic reference-free evaluation is cheap and fast, it may be unreliable. Toxicity and fluency classifiers are not perfect and can return erroneous evaluations. The embedding distance which is used to measure the content preservation was shown to weakly correlate with human judgements~\cite{yamshchikov2021}. Thus, we evaluate the best-performing models manually. % to get a more reliable information on their performance.

%The classifiers for evaluating toxicity and fluency can make mistakes, so their judgements cannot be considered ground truth. Besides that, the CoLA dataset used for training the fluency classifier contains human-written sentences. Thus, it has examples of  human disfluencies and mistakes, which is not completely relevant for the evaluation of machine-generated text. Finally, the evaluation of content preservation is conducted via embedding distances which were shown to have a weak correlation with human judgements~\cite{}. Besides that, 

\paragraph{Experimental Setup}

We design our manual evaluation setup to be as close as possible to the automatic evaluation. We evaluate our models along the same three metrics: style accuracy (ACC$_m$), content similarity (SIM$_m$), and fluency (FL$_m$). For all metrics we use a ternary scale: \{0, 0.5, 1\} corresponding to a bad, partially acceptable, and fully acceptable sentence. %Our criteria of acceptability are not rigorous. We do not label a text as toxic if it contains passive aggression and covert toxicity which can be interpreted in a non-toxic way. Likewise, we allow minor changes of content and minor grammatical and punctuation mistakes which do not distort the text sense.

We ask five annotators to evaluate the models. Annotators are NLP researchers with MSc degree or above and with a good command of English. Prior to the annotation, we arranged a preliminary round to reach common annotation understanding. Each sentence is evaluated by three annotators, the final score for a sentence is computed as the average of their scores. We measure the inter-annotator agreement in terms of Krippendorff's $\alpha$. We obtained the score of 0.42 for the style accuracy, 0.31 for content preservation, and 0.52 for fluency: a moderate agreement for style and fluency annotation, and low agreement for content annotation.

We evaluate three models: our new models ParaGeDi and CondBERT, and Mask\&Infill whose automatic scores were the highest among the existing models.
The evaluation was conducted on 200 source sentences, each of them was transformed by each of the evaluated models. The input (toxic) sentences for the evaluation were manually pre-selected to filter out disfluent or senseless utterances (this pre-selection did not consider the outputs). To compensate for the low inter-annotator agreement, we annotate each sample three times and report the average score.

%Our preliminary experiments showed that disagreements in annotation often stem from the fact that the original sentences are disfluent or senseless. Thus, their detoxified versions are difficult to interpret. In order to make the manual evaluation more reasonable and useful, we manually pre-select 200 source (toxic) sentences for the evaluation. We emphasise that we perform this selection solely based on the original sentences and do not consider their rewritten versions. Therefore, this pre-selection does not give any preference to any of the models.

\paragraph{Discussion of Results}

We show the performance of models in terms of human evaluation in Table~\ref{tab:manual_result}. The model scores are the averaged sentence scores. We combine the three metrics into a joint quality score which we denote as \textbf{J}$_{\textbf{m}}$. % (\underline{\bf J}oint metric for \underline{\bf m}anual evaluation). 
Sentence-level \textbf{J}$_{\textbf{m}}$ is a multiplication of sentence ACC$_m$, SIM$_m$, and FL$_m$, and the model \textbf{J}$_{\textbf{m}}$ scores are the average of sentence scores. %Given the ternary nature of all metrics, the \textbf{J}$_{\textbf{m}}$ of 1 is given to sentences which are fully acceptable in terms of all three metrics. Conversely, a score of 0 along any of the metrics turns overall \textbf{J}$_{\textbf{m}}$ to 0. Therefore, \textbf{J}$_{\textbf{m}}$ is a rough estimate of the percentage of acceptable detoxified sentences generated by a model.
This manual evaluation corroborates the superiority of our models over Mask\&Infill model. At the same time, it confirms that our two models are not significantly different. Although ParaGeDi outperforms CondBERT in terms of all metrics, the difference in scores is statistically significant only for FL$_m$. 

%The differences between ParaGeDi and Mask\&Infill and between CondBERT and Mask\&Infill are significant.
%Besides the discussed metrics, we report the percentage of good sentences (rightmost column of Table~\ref{tab:manual_result}). These are the percentages of sentences which got the scores of 1 for all metrics from all annotators. This is the estimate of the percentage of answers from a model which would be acceptable in a real-world application. 

% \begin{table}[]
% \centering
% \footnotesize
% \begin{tabular}{l|l|l|l|l|l}
% \toprule
%                                                                       & \multicolumn{1}{l}{ACC} & \multicolumn{1}{l}{SIM} & \multicolumn{1}{l}{FL} & \multicolumn{1}{l}{\textbf{J}$_{\textbf{m}}$} & \% good \\ \midrule
% \rowcolor{green!30} ParaGeDi  & \textbf{93.41} & \textbf{64.75}  & \textbf{91.25}* & \textbf{55.34} & \textbf{20.5*}\\
% \rowcolor{green!30} CondBERT  & 91.00*  & 63.92* & 86.41* & 50.47*  & 14.0*\\
% Mask\&Infill                  & 75.33  & 59.08 & 62.08 & 27.33 & 1.5
% \\ \bottomrule                
% \end{tabular}
% \caption{The results of manual evaluation sorted with respect to \textbf{J}$_{\textbf{m}}$. The highest scores are shown \textbf{in bold}. The differences in scores from the next best model marked by * are statistically significant with $\alpha<0.05$. The significance was evaluated with the paired \textit{t-test}.}
% \label{tab:manual_result}
% \end{table}

\begin{table}
\centering
\footnotesize
\begin{tabular}{l|l|l|l|l}
\toprule
                                                                      & \multicolumn{1}{l}{\textbf{ACC}$_m$} & \multicolumn{1}{l}{\textbf{SIM}$_m$} & \multicolumn{1}{l}{\textbf{FL}$_m$} & \multicolumn{1}{l}{\textbf{J}$_{\textbf{m}}$}  \\ \midrule
\rowcolor{green!30} ParaGeDi (ours)  & \textbf{93.41} & \textbf{64.75}  & \textbf{91.25} & \textbf{55.34} \\
\rowcolor{green!30} CondBERT (ours)  & 91.00  & 63.92 & 86.41 & 50.47  \\
Mask\&Infill (top 1)                & 75.33  & 59.08 & 62.08 & 27.33 
\\ \bottomrule                
\end{tabular}
%\caption{The results of manual evaluation sorted by \textbf{J}$_{\textbf{m}}$. The differences in scores from ParaGEDI/CondBERT to the top 1 baseline, according to the automatic evaluation are statistically significant with $\alpha<0.05$ based on the paired \textit{t}-test. Differences between ParaGeDi and CondBERT are significant only for the FL$_m$ metric.}
\caption{The results of manual evaluation sorted by \textbf{J}$_{\textbf{m}}$. The differences between our models and Mask\&Infill are statistically significant with $\alpha<0.05$ based on the paired \textit{t}-test. Differences between ParaGeDi and CondBERT are significant only for the FL$_m$ metric.}
\label{tab:manual_result}
\end{table}

\begin{table*}[ht!]
\centering
\footnotesize
\begin{tabular}{l|rrrr|r}
\toprule
% {} & Style & \multicolumn{1}{c|}{Similarity} & \multicolumn{2}{c|}{Naturalness} & \multicolumn{2}{c|}{Aggregate} & \\
 %\midrule
 \textbf{Model}       & \bf ACC & \bf  SIM & \bf  FL & \bf  J & \bf  BLEU \\
 \midrule
%original text    & 0.02  & 1.00  & 104 & 0.87 & 0.052 & 0.015 & 0.180 \\
human            & 0.81  & 0.65  & 0.84 & 0.445 $\pm$ 0.011~~ & 1.000  \\
\midrule
\rowcolor{green!30}ParaGeDi (ours)         & 0.93  & 0.62  & 0.88 & \textbf{0.515} $\pm$ 0.009* & 0.038 $\pm$ 0.005 \\
Mask \& Infill \cite{MLM1} & 0.89  & 0.76  & 0.62 & 0.420 $\pm$ 0.013~~ & 0.145 $\pm$ 0.008 \\
DualRL \cite{luo2019dual}& 0.87  & 0.75 & 0.63 & 0.395 $\pm$ 0.012~~ & \textbf{0.152} $\pm$ 0.008 \\
\rowcolor{green!30}
CondBERT (ours) & 0.86  & 0.65  & 0.62 & 0.338 $\pm$ 0.012~~ & 0.125 $\pm$ 0.007 \\
%UnsuperMT\_Zhang& 0.96  & 0.64 & 256 & 0.47 & 0.134 & 0.29 & 0.109 \\
SST \cite{lee2020stable}& 0.74  & 0.65  & 0.41 & 0.225 $\pm$ 0.011~~ & 0.100  $\pm$ 0.007 \\
DRG-RetrieveOnly \cite{DRG} & 0.95  & 0.29  & 0.83 & 0.225 $\pm$ 0.006~~ & 0.004 $\pm$ 0.001  \\
DRG-TemplateBased \cite{DRG} & 0.82  & 0.70  & 0.24 & 0.115 $\pm$ 0.009~~ & 0.117 $\pm$ 0.007 \\
 \bottomrule
\end{tabular}
\caption{Performance of the sentiment transfer models on the YELP dataset. The models are sorted with respect to the aggregated \textbf{J} score. * indicates the score which is significantly higher than the next best model with $p<0.01$.}
\label{table:sentiment}
\end{table*}

\begin{table}
\footnotesize
\centering
\begin{tabular}{ll|ll|ll}
\toprule
\multicolumn{2}{c|}{\bf ACC$_m$} & \multicolumn{2}{c|}{\bf SIM$_m$} & \multicolumn{2}{c}{\bf FL$_m$} \\ \midrule
ACC-soft           & 0.59      & SIM          & 0.34         & FL           & 0.54         \\
ACC      & 0.51      &  BLEU         & 0.19        & PPL          & 0.45         \\
\bottomrule
\end{tabular}
\caption{Spearman's $\rho$  of automatic metrics for evaluating style, content, and fluency with our human scores.}
\label{tab:manual_correlation}
\end{table}

%The goal of our manual evaluation is twofold. 
Besides the evaluation itself, we investigated to what extent the automatic metrics reflect the human judgements. To do that, we compute their Spearman's $\rho$ correlation score with human judgements (see Table~\ref{tab:manual_correlation}). %For each of the evaluation dimensions (style, content, and fluency), we compute the Spearman's $\rho$ score of our manual scores with the values of automatic metrics. 
For style, we consider the accuracy of toxicity classifier that we used for the evaluation (ACC) and its version which returns the confidence instead of the binary label (ACC-soft). For content we compare SIM (embedding similarity used for computing the \textbf{J} score) and BLEU score between the original and detoxified sentence. For fluency, we consider the linguistic acceptability classifier (FL) and perplexity of the GPT-2 \cite{gpt2} language model (PPL) which is used for evaluating fluency in many works on style transfer and other generation tasks.

This evaluation shows that the tested metrics of content preservation show only weak correlation with manual scores, which agrees with the previous research~\cite{yamshchikov2021}. The correlation of automatic style and fluency metrics with human judgements is moderate. It turns out that the confidence of style classifier is a better style accuracy metric than a binary classifier and the acceptability classifier works better than perplexity, confirming the criticism of perplexity as a fluency metric~\cite{krishna-etal-2020-reformulating}.

\section{Sentiment Transfer Experiments}
%label{app:sentiment}

Text detoxification is not as well-established as other style transfer tasks, which makes it is difficult to put our models in the context of other works on style transfer. Thus, we conduct an experiment on a different domain, namely, sentiment transfer.

\textbf{Experimental Setup}~~We train ParaGeDi and CondBERT on the Yelp reviews dataset \cite{DRG} and compare them with Mask\&Infill, SST, DRG-TemplateBased, DRG-RetrieveOnly, and DualRL models (see Section~\ref{section:related}). 
%~\cite{luo2019dual} model which solves dual tasks of transfer from style $X$ to $Y$ and back from $Y$ to $X$ and uses the feedback of the models to each other and the style classifiers as a signal. 
We tune the hyperparameters of ParaGeDi and CondBERT on the Yelp development set and use the outputs of other models generated by their authors. 

We evaluate the models using the \textbf{J} as in our detoxification experiments. For the evaluation of style transfer accuracy, we train two sentiment classifiers on two disjoint parts of the Yelp dataset as in Section \ref{section:clf}. We use one for inference and another for evaluation. We also compute the BLEU score against human references provided by \newcite{DRG}.
The results are shown in Table \ref{table:sentiment}, averaged over two transfer directions.

\textbf{Discussion of Results}~~ParaGedi outperforms other models in terms of \textbf{J}. % due to better style accuracy and higher fluency. 
As before, the other models fail to generate fluent texts because they replace only specific words or because they learn to generate texts from scratch. ParaGeDi model is the only competitor which combines pre-trained models and with full regeneration. The performance of the CondBERT model is low on this task, corroborating that detoxification and style transfer for other domains require different techniques.

On the other hand, the BLEU score questions this result. Compared to the human references, the best-performing model is \textbf{DualRL} followed by the two MLM-based models: Mask\&Infill and our CondBERT.  
The evaluation of reference human answers also questions the referenceless metrics. First, we see that the ACC score is limited by the classifier performance. Since it gives only 0.81 to presumably 100\% correct manually written sentences, the small differences in ACC should not be considered significant, and the ACC above 0.81 is unreliable. Overall, since the score of human answers is close to those of ParaGeDi and Mask\&Infill, ParaGeDi can still be considered a strong style transfer model, and more precise evaluation should be done by humans because metrics cannot distinguish between the models at this level.

\section{Conclusion}

We present two style transfer models tailored for detoxification, i.e. transfer from toxic to non-toxic texts. Both of them combine high-quality pre-trained LMs with the extra style guidance. \textbf{ParaGeDi} is based on a paraphraser guided by a style-conditioned GPT-2 model. \textbf{CondBERT} model is based on BERT which does not need any fine-tuning, and all style control is performed with a pre-trained toxicity classifier.
We conduct a large-scale study of style transfer models exploiting both automatic and manual evaluation. 
Our experiments show that the proposed methods outperform other state-of-the-art style transfer models on the tasks of detoxification and sentiment transfer.

%We suggest that future work on detoxification should focus on (i) the use of pre-trained models and (ii) development of evaluation metrics better correlated with human judgement. 

\section*{Ethical Statement}

Toxicity is a sensitive topic where the unexpected results and byproducts of research can cause harm. Therefore, we would like to consider some ethical concerns related to our work.

\paragraph{On Definition of Toxicity} 

Toxicity is an umbrella term for almost any undesirable behaviour on the Internet. It ranges from ``mild'' phenomena like condescending language~\cite{perez-almendros-etal-2020-dont} to grave insults or oppression based on racial or other social-demographic characteristics. 

While annotators agree when labelling serious cases of toxicity such as hate speech~\cite{Fortuna2018}, the labelling of less severe toxicity is subjective and depends on the annotator's background~\cite{al-kuwatly-etal-2020-identifying}. This can cause the underestimation of certain types of toxicity. To define the toxicity in the most objective feasible way, we adopt a data-driven approach as presented in detail formally in Appendix~\ref{sec:definition}. Both models we propose recognise toxicity based on a toxicity-labelled dataset and do not require any additional manually created dictionaries or rules. Thus, their understanding of toxicity can be tuned with the input data. This ensures that given a corpus with unbiased toxicity labelling our models can produce unbiased detoxification.

On the other hand, in case the training corpus is biased, the model can reproduce the biases, so it should be applied with caution.

\paragraph{Toxification of Texts}

Detoxification task implies the possibility to perform the opposite transformation, i.e. to rewrite a neutral text into a toxic one. Various style transfer models, including ours, could in principle be used to complete this task. However, in case of CondBERT, the quality of such transformation would be bad, and it would be almost impossible to pass the results of this ``toxification'' off as real toxic sentences. The reason for that is the structure of toxic data.

One of the main properties of toxic style is the presence of lexical markers of this style (rude or obscene words). Such markers (i) carry most of stylistic information of a sentence (i.e. their presence is a strong indicator of this class), (ii) have synonyms which are free from this stylistic information. Both our methods strongly rely on these properties. They identify toxic words and replace them with non-toxic synonyms. On the other hand, if performing the opposite transformation, we cannot use these properties any more. First, there do not exist non-toxic words which are strong indicators of neutral (non-toxic) style. Second, it is almost infeasible to identify non-toxic words which have toxic synonyms and replace them appropriately. Therefore, we suggest that CondBERT is not suitable for toxification.

The above arguments do not prove that CondBERT or ParaGeDi cannot be applied for toxification. However, they suggest that the quality of the resulting text might not be higher than with simpler toxification methods (e.g. handwritten rules for inserting rude words).

\paragraph{Detoxification as a Censorship}

Another concern is the fact the detoxification technology could no used to rewrite user-generated messages, which might be considered a form of censorship. We would like to look at that from a different perspective. The social media currently already perform censorship, e.g. Instagram provides tools for removal of messages based on automatically identified harmful content.\footnote{\href{https://about.instagram.com/blog/announcements/introducing-new-tools-to-protect-our-community-from-abuse}{https://about.instagram.com/blog/announcements/introducing-new-tools-to-protect-our-community-from-abuse}}

On the other hand, we suggest mitigating this policy by rewriting toxic messages instead of removing them altogether. Last but not least, we suggest that user messages should not be modified without user consent. The detoxificaiton models should be used for suggesting detoxifying edits rather than performing them automaticallly.

At the same time, detoxification models can make chatbots safer by detoxifying (if necessary) their answers before sending them to users. An automatically generated toxic comment by a neural chatbot may be the result of pre-training on the biased textual data -- a problem which is currently unsolved completely~\cite{gehman-etal-2020-realtoxicityprompts}. Therefore, a detoxification of automatically generated content might be a valid use-case for minimizing reputational losses for the company created such an unmoderated chatbot~\cite{babakov-etal-2021-detecting}. 

\section*{Acknowledgements}

This research was conducted under the framework of the joint MTS-Skoltech laboratory. We are grateful to the reviewers for their helpful suggestions which substantially improved this work.

\bibliography{anthology,aacl-ijcnlp2020}
\bibliographystyle{acl_natbib}

\clearpage

\appendix

\section{Definition of Text Detoxification Task}
\label{sec:definition}

In our work, we adhere to the data-driven definition of toxicity. The toxicity is a particular binary value associated with a text: \{\texttt{toxic}, \texttt{neutral}\}. We assume that this textual characteristic is measurable using a function $\sigma(x_i) \rightarrow s_i$ that obtains as input text $x_i$ and returns the corresponding style label $s_i$. For instance, it can be implemented using a \mbox{text classifier.}

Let us assume a set of two discreet mutually exclusive styles $S = \{ s^{src}, s^{tg} \}$ which corresponds to the source \texttt{toxic} and target \texttt{neutral} styles. Let us consider two text corpora ${D^{src}} = \{d_1^{src}, d_2^{src}, ..., d_n^{src}\}$ and ${D^{tg}} = \{d_1^{tg}, d_2^{tg}, ..., d_m^{tg}\}$ belonging to the source and target styles $s^{src}$ and $s^{tg}$, respectively. For each text $d_i$, let us assume that it has a style $s_i$ measurable with the function $\sigma: D \rightarrow S$. There also exists a binary function $\delta: D \times D \rightarrow [0, 1]$ that indicates the semantic similarity %degree of equivalence of meaning 
of two input texts and a unary function $\psi: D \rightarrow [0, 1]$ that indicates the degree of the text fluency. In general, the sizes of the source and the target corpora ${D^{src}}$ and ${D^{tg}}$ are different ($n \neq m$) and the texts in them are not aligned, i.e., in general, $\delta(d_i^{src}, d_i^{tg}) \neq 1$. If $n=m$ and $\delta(d_i^{src}, d_i^{tg}) = 1$ for all texts, this is a special case of a parallel style-aligned corpus. Given the introduced notations, we define the task of text detoxification as follows:

A text detoxification model is a function $\alpha:~S~\times~S~\times~D~\rightarrow~D$ that, given a source style $s^{src}$, a target style $s^{tg}$, and an input text $d^{src}$, produces an output text $d^{tg}$ \mbox{such that:}
\begin{itemize}

\item The style of the text changes from the source style $s^{src}$ to the target style \mbox{$s^{tg}$: $\sigma(d^{src}) \neq \sigma(d^{tg})$,} $\sigma(d^{tg}) = s^{tg}$;
\item The content of the source text is saved in the target text as much as required for the task: $\delta(d^{src}, d^{tg}) \geq t^{\delta}$;
\item The fluency of the target text achieves the required level: $\psi(d^{tg}) \geq t^{\psi}$,

\end{itemize}
where $t^{\delta}$ and $t^{\psi}$ are the error tolerance threshold values for the content preservation ($\delta$) and fluency ($\psi$) functions. 
 
When removing the toxicity from a text, we inevitably change a part of its meaning, so full content preservation cannot be reached. However, we should attempt to save the content as much as possible and adjust $t^{\delta}$ to the needs of this task. 

Thus, the task of obtaining a text detoxification model with the best parameters set may be viewed as maximizing the probability $P(d^{tg}|d^{src},s^{src}, s^{tg})$ given the three above-mentioned constraints based on parallel or non-parallel text corpora $D^{src}$ and $D^{tg}$.

\section{CondBERT Ablation Study}
\label{appendix:condbert}

Our CondBERT model was inspired by \newcite{condbert} and is similar to \newcite{MLM1}, but has some unique properties. We test their importance with the ablation study on the detoxification task. 

\begin{table}[ht!]
\centering
\small
\begin{tabular}{l|r|r|r|r}
\toprule

 \bf  Model       & \bf ACC  & \bf SIM &\bf   FL & \bf J \\
 \midrule
Full model               & 0.91  & 0.73  & 0.75 & 0.50  \\
 \midrule

Mask all toxic tokens    & 0.94  & 0.69  & 0.77 & 0.50  \\
No similarity penalty    & 0.92  & 0.73  & 0.75 & 0.50 \\
No multiword replacement              & 0.87  & 0.80  & 0.56 & 0.40 \\
No toxicity penalty      & 0.57  & 0.79  & 0.82 & 0.33 \\

 \bottomrule
\end{tabular}
\caption{Results of the CondBERT ablation study on detoxification.}
\label{table:condbert_ablation}
\end{table}

We use two heuristics for content preservation: not masking the toxic tokens and reranking replacement candidates with respect to their similarity to the original tokens. Removing any of these heuristics leads to lower content preservation and higher style accuracy, showing the inverse correlation of these properties (see Table~\ref{table:condbert_ablation}). However, the \textbf{J} score for these models stays the same. 
On the other hand, turning off the possibility of filling a single mask with multiple words reduces the fluency and style accuracy, although obviously yields a better content preservation score, because the output sentence contains less new words. This affects the \textbf{J} score which is reduced for this model. Finally, the greatest impact on the \textbf{J} metric is caused by eliminating the toxicity penalty. The ACC is reduced dramatically, and although the other two metrics slightly grow compared to the full model, they cannot compensate for the low style accuracy.

\section{ParaGeDi Ablation Study}
\label{appendix:paragedi}

ParaGeDi model consist of a paraphraser, a language model trained with the generative-discriminative loss, and a style classifier for reranking hypotheses. In addition to that, we use a number of heuristics during inference. We conduct ablation study on the detoxification task to understand the usefulness of these components. We test the following variations of ParaGeDi:

\begin{itemize}
    \item no discriminative loss ($\lambda=0$),
    \item no generative loss ($\lambda=1$),
    \item no upper bound ($u=\infty$),
    %\item no lower bound ($l=-\infty$),
    \item no discriminator ($w=0$),
    \item no extra control of style strength ($w=1$),
    \item no probability smoothing ($\alpha=0$),
    \item no reranking,
    \item no beam search (beam size of 1).
\end{itemize}

In each configuration, all other parameters are fixed.
The performance of models is given in Table~\ref{table:gedi_ablation}. 

Decreasing the number of beams leads to the deterioration of fluency and of style strength because of the smaller number options for the reranker to choose from. Removing the reranker leads to lower style strength with small gains in similarity or fluency. Turning off the smoothing of probabilities makes similarity and fluency degrade a little. Removing the upper bound on the discriminative correction leads to nearly 100\% style transfer but to very low similarity of the generated sentences to the original ones, as the model starts hallucinating. Decreasing the $w$ parameter reduces style accuracy but improves fluency and similarity, showing a clear trade-off between them.

The individual components of the loss are slightly less important for style than inference parameters. With only the discriminative loss the model is still able to successfully transform style in 77\% of the cases, and the generative loss alone is able to change the style in 90\% cases. The latter figure shows that the model equipped with style labels can discriminate between styles even if it was not explicitly trained to do that. On the other hand, the elimination of the generative loss results in a significant drop in fluency. Although the class-conditional LM in ParaGeDi is a GPT2 model which has already been trained for generation task, the lack of generation-based fine-tuning reduces the quality of the resulting text.

\begin{table}
\centering
\small
\begin{tabular}{l|r|r|r|r}
\toprule
 \bf  Model       & \bf  ACC & \bf  SIM & \bf  FL & \bf  J\\
 \midrule
 
Full ParaGeDi             & 0.95 & 0.66 & 0.79 & 0.50 \\
\midrule
%$w=1$ \\
%$w=0$ \\
no reranking              & 0.87 & 0.70 & 0.80 & 0.50 \\
beam of size 5            & 0.93 & 0.67 & 0.79 & 0.50 \\
no discriminative loss    & 0.90 & 0.68 & 0.81 & 0.49 \\
no smoothing              & 0.96 & 0.63 & 0.78 & 0.47 \\
no beam search            & 0.88 & 0.66 & 0.70 & 0.41 \\
no control of style strength &0.56& 0.80 & 0.81 & 0.38 \\
no generative loss        & 0.77 & 0.70 & 0.67 & 0.37 \\
no upper bound            & 0.99 & 0.35 & 0.76 & 0.27 \\
\bottomrule
\end{tabular}
\caption{Results of the ParaGeDi ablation study.}
\label{table:gedi_ablation}
\end{table}

\section{Details of Mining the Parallel Corpus}
\label{appendix:parallel}

Here we describe in more detail the process of mining of a detoxifying parallel paraphrase corpus. We use the ParaNMT dataset \cite{wieting2017paranmt} comprised of 50 million English sentences and their paraphrases back-translated from Czech. We filter the dataset keeping only the sentence pairs with moderate similarity (0.6 to 0.95) and similar length (with difference at most 40\%), which is approximately 50\% of the dataset. We compute the similarity as the cosine distance between the averaged BERT embeddings of all words in a sentence. After this similarity- and length-based filtering we score each sentence with a RoBERTa-based toxicity classifier from Section~5.1 and keep only the pairs with the difference in toxicity scores of at least 50\%. Thus, we obtain 500,000 sentence pairs. Their examples are given in Table \ref{table:mined_examples}.

Manual inspection of a random sample of the selected pairs shows that around 10\% of them are invalid paraphrases, 40\% are in fact both toxic or both safe, and around 50\% of them are valid detoxifying paraphrases. This suggests that with more rigorous filtering we can yield a corpus for detoxification of around 250,000 high-quality parallel sentences, which is larger than the majority of existing parallel style transfer datasets.

% \usepackage[normalem]{ulem}
\begin{table*}[ht]
\small
\begin{tabular}{c p{0.33\linewidth}  p{0.33\linewidth} llll}
\toprule
\textbf{} &
  \multicolumn{1}{c}{\textbf{Reference}} &
  \multicolumn{1}{c}{\textbf{Translation}} &
  \multicolumn{1}{c}{\textbf{sim}} &
  \multicolumn{1}{c}{\textbf{ld}} &
  \multicolumn{1}{c}{\textbf{tox\textsubscript{ref}}} &
  \multicolumn{1}{c}{\textbf{tox\textsubscript{trn}}} \\
\midrule
\textbf{0} &
  If   Alkar is flooding her with psychic waste, that explains the high level of   neurotransmitters. &
  if   Alkar floods her with her mental waste, it would explain the high levels of   neurotransmitter. &
  0.78 &
  0.01 &
  0.01 &
  0.98 \\
\textbf{1} &
  Now   you're getting nasty. &
  you're   becoming disgusting. &
  0.75 &
  0.07 &
  0.06 &
  0.99 \\
\textbf{2} &
  Well,   we could spare your life, for one. &
  well,   we can spare your life. &
  0.92 &
  0.27 &
  0.21 &
  0.98 \\
\textbf{3} &
  Ah!   Monkey, you've got to snap out of it. &
  monkey,   you have to wake up. &
  0.66 &
  0.31 &
  0.05 &
  0.99 \\
\textbf{4} &
  I've   got orders to put her down. &
  I   have orders to kill her. &
  0.73 &
  0.18 &
  0.01 &
  0.99 \\
\textbf{5} &
  I'm   not gonna have a child... ...with the same genetic disorder as me who's gonna   die. L... &
  I'm   not going to breed kids with a genetic disorder that makes them die. &
  0.70 &
  0.20 &
  0.95 &
  0.04 \\
\textbf{6} &
  They're   all laughing at us, so we'll kick your ass. &
  they're   laughing at us. We'll show you. &
  0.62 &
  0.23 &
  0.99 &
  0.00 \\
\textbf{7} &
  Maine   was very short on black people back then. &
  there   wasn't much black in Maine then. &
  0.72 &
  0.19 &
  0.96 &
  0.15 \\
\textbf{8} &
  Briggs,   what the hell's happening? &
  Briggs,   what the hell is going on? &
  0.92 &
  0.00 &
  0.16 &
  0.84 \\
\textbf{9} &
  Another   one simply had no clue what to do, so whenever he met my brother he'd beat   the crap out of him, and then say: &
  another   simply didn't know what to do, so whenever he met my brother, he nearly beat   the shit out of him. &
  0.88 &
  0.10 &
  0.05 &
  0.93 \\
\textbf{10} &
  I   suppose you want me to buy you flowers and chocolates and whisper sweet   nothings. &
  you'd   probably want me to buy you some chocolates and flowers... and whispered some   pretty rubbish. &
  0.80 &
  0.16 &
  0.00 &
  0.98 \\
\textbf{11} &
  So   now their spirits are cursed, walking back roads, waterways, and if they find   an unfaithful man, they kill him, and that man is never seen again. &
  their   souls are cursed, they guard the paths, he says, and when they encounter an   unfaithful man, he will be killed, and his body will never be found. &
  0.75 &
  0.01 &
  0.84 &
  0.14 \\
\textbf{12} &
  Freezing   him. &
  I'll   freeze him! &
  0.78 &
  0.18 &
  0.01 &
  0.57 \\
\textbf{13} &
  Come   on, Cal, leave that shit alone. &
  come   on, Cal, put it down. &
  0.66 &
  0.27 &
  0.99 &
  0.00 \\
\textbf{14} &
  So   he's the Top dog. &
  he's   the tallest son of a bitch. &
  0.61 &
  0.36 &
  0.00 &
  0.99 \\
\textbf{15} &
  I   swore when I went out with Xander Harris... ...I'd rather die than datea   fixer-upper again. &
  when   I was dating Alex Harris, I swore I'd rather die than go out with a loser. &
  0.79 &
  0.15 &
  0.01 &
  0.99 \\
\textbf{16} &
  I'm   famous, and you're done. &
  I'm   famous, and you're dead. &
  0.82 &
  0.00 &
  0.00 &
  0.98 \\
\textbf{17} &
  To   quote Jake Oppenheimer: I, who am about to die, must seem to them something   ``God-awful.''... &
  to   quote Jake and Oppenheimer: ``I must die, I must feel like a terrible   god.'' &
  0.70 &
  0.18 &
  0.00 &
  0.68 \\
\textbf{18} &
  “Could   you please be quiet, Miss Lavish?” said Moist. &
  'could   you keep your mouth shut, Miss Opulent? 'Said Moist. &
  0.81 &
  0.10 &
  0.00 &
  0.76 \\
\textbf{19} &
  Murder   for hire. &
  murder   to order. &
  0.70 &
  0.00 &
  0.07 &
  0.96\\
\bottomrule
  
\end{tabular}
\caption{Examples of mined detoxifying paraphrases. Here \textbf{sim} is similarity between sentences, computed by  \cite{wieting2017paranmt}, \textbf{ld} is relative difference in length, \textbf{tox}\textsubscript{ref} and \textbf{tox}\textsubscript{trn} are toxicity scores calculated by our classifier.}
\label{table:mined_examples}
\end{table*}

\section{Qualitative Analysis}
\label{sec:qualitative}

Both automatic and manual joint scores show that our best models are halfway between useless and perfect. But the actual success rate is much less than half. We call a detoxified sentence ``perfect'' if all three annotators gave it the maximal scores for all three aspects. With this definition, only 20\% of ParaGeDi sentences and 14\% of CondBERT sentences are perfect, and only about 1.5\% of Mask\&Infill sentences are perfect. %These low success rates suggest that our current models should not be used in automatic mode, and should only suggest possible rewrites that a user can accept or deny.

As you can judge from Table~\ref{tab:manual_result}, the main cause of imperfection for all models is distortion of meaning. Below we describe our manual investigation into the causes of this distortion.

In half of the cases, ParaGeDi conveys the meaning more or less adequately. 
Its mistakes include:

\begin{itemize}

\item replacement of toxic words by similarly looking less toxic words with different meaning (e.g. ``whore''  $\rightarrow$ ``Who's who'', ``stop behaving like fascist thugs'' $\rightarrow$ ``Stop looking at fascism'', ``taxman massive cunt , only outdone by linuxcunt himself .'' $\rightarrow$ ``Taxman's massive cut, outdone by Linuxcune himself.'').

\item replacement of sentence meaning (``the election was yours to lose'' $\rightarrow$ ``the election is to be won'', ``this crap hole institute run by motherfuckers and bastards'' $\rightarrow$ ``a deloitte institute for mothers and their children'')

\item Avoiding the toxic or difficult part, for example ``why we gotta have this miscegenation crap ?'' $\rightarrow$ ``Why do we need to have it?''. 
In some cases, however, ParaGeDi masks or rephrases the toxic part of the message, while still preserving the general meaning, for example ``start there first you idiot!'' $\rightarrow$ ``Let's start there first.''
\end{itemize}

In general, ParaGeDi makes the impression of fantasising too much, because it often rewrites the whole sentence, and from time to time changes its structure significantly.

CondBERT, on the other hand, usually preserves the sentence structure, but often replaces words with inappropriate substitutes, often antonyms: ``selfish'' $\rightarrow$ ``misunderstood'', ``racists'' $\rightarrow$ ``politicians'', ``cunt'' $\rightarrow$ ``nursemaid'', ``to troll and harass'' $\rightarrow$ ``to try out and help'',  ``asskissers'' $\rightarrow$ ``honest people'', ``retarded'' $\rightarrow$ ``beautiful'', ``whore'' $\rightarrow$ ``sweetheart''. Sometimes these replacements are more adequate, e.g. ``old cock'' $\rightarrow$ ``old-fashioned stuff'', ``your attitude is shit'' $\rightarrow$ ``your attitude is completely wrong'', ``bitch i warned yall'' $\rightarrow$ ``i guess i warned you''. But the proportion of sentences with the meaning preserved is smaller for CondBERT, being less than 40\%.

The DLSM and Template-based DRG models often preserve the meaning by just preserving the toxic words, so their total success rate is low. The Retrieve-only DRG model almost never preserves the meaning. The Mask\&Infill model seems to be overfitted: it often replaces toxic words with irrelevant non-toxic words (e.g. ``crap'' $\rightarrow$ ``compelling'') that the model apparently considers to be the ``markers'' of the non-toxic style. These properties make the baselines unsuitable for the detoxification task without adaptation, and the CondBERT model is in fact such an adaptation. %As the reviewers note, CondBERT is an incremental change to the existing style transfer approach, and this is our deliberate decision: with this model, we investigate how suitable are the existing approaches for the detoxification problem if we apply only minimal adaptations to them.

Typical mistakes of both ParaGeDi and CondBERT can be attributed mostly to the insufficiency of semantic understanding: they often replace toxic words with semantically related words of different (often opposite) meaning, or simply with similarly looking words. We conjecture that with a paraphraser trained on a larger corpus (we have used only  2\% of ParaNMT) or on more difficult examples would improve the ability of  ParaGeDi to preserve meaning.

Generally, our models produce the impression of not being mature enough for fully automatic use with texts where meaning is important. However, they can be used to suggest detoxification options to human writers, or to detoxify the output of chit-chat bots where the cost of producing an inarticulate utterance is considerably less than the cost of producing a toxic one.

\section{Examples of Detoxification}
\label{appendix:detox_examples}

Table~\ref{table:detox_examples_appendix} shows the examples of detoxification by our models ParaGeDi and CondBERT and the best competitor Mask\&Infill.

\begin{table*}[htb]
\centering
\footnotesize
\begin{tabular}{p{0.15\linewidth}  p{0.85\linewidth}}
\toprule
Input &  id have relatives signing up for aol finally in the mid to late 90s forwarding me stupid conservative conspiracy theory garbage . \\
ParaGeDi (ours) & I'd have relatives sign up for aol in the mid-to late 90s. and send me some unorthodox conspiracy theories. \\
CondBERT (ours) & id have relatives signing up for aol finally in the mid to late 90s forwarding me with some very conservative conspiracy theory stuff . \\ 
Mask\&Infill & id have relatives signing up for aol finally in the mid to late 90s forwarding me to conservative conspiracy theory ideas .	 \\
\midrule
Input & why dont you bung off . smoke a fag . and have a cuppa ?     \\
ParaGeDi (ours) & Why don't you take a break. smoke a cigar. and have a cup of tea? \\
CondBERT (ours) & why dont you bung off . smoke a great deal . and have a cuppa ? \\
Mask\&Infill & why dont you bung off . smoke a compelling compelling . and have a cuppa ? \\
\midrule
Input & can someone please fix that crap ? \\
ParaGeDi (ours) & Can someone please do that? \\
CondBERT (ours) & can someone please fix that stuff ?	 \\ 
Mask\&Infill & can someone please fix that compelling ? \\
\midrule
Input & its a very stupid name ! \\
ParaGeDi (ours) & It's a very strange name! \\
CondBERT (ours) & its a very . . . strange . . . name ! \\ 
Mask\&Infill & its a very compelling name ! \\
\midrule
Input & im a rookie pilot and would have never have done something this stupid . \\
ParaGeDi (ours) & I've been a pilot for a while. and I never would have done anything like \\
CondBERT (ours) & im a rookie pilot and would have never have done something this wrong .	 \\ 
Mask\&Infill & im a rookie pilot and would have never have done something this compelling . \\
\midrule
Input	&	cut the damn funding and lets move on to the next program .	 \\
ParaGeDi (ours) &	Cut the funding and move on to the next program. \\
CondBERT (ours) & 	cut the amount of the funding and lets move on to the next program . \\
Mask\&Infill &	cut the compelling funding and lets move on to the next program . \\
\midrule
Input	&	bitch i warned yall . didnt i ?	 \\
ParaGeDi (ours) &	Bits I've warned a lot. right? \\
CondBERT (ours) & 	i guess i warned you . didnt i ? \\
Mask\&Infill &	compelling i warned yall . didnt i ? \\
\midrule
Input	&	i was such an asshole .	 \\
ParaGeDi (ours) &	I was an ad. \\
CondBERT (ours) & 	i was such an emotional mess . \\
Mask\&Infill &	i was such an compelling . \\
\midrule
Input&		your statements show me how ignorant you are .	 \\
ParaGeDi (ours) &	Your statements show me that you'd be indifferent. \\
CondBERT (ours) & 	your statements show me how very wronged you are . \\
Mask\&Infill &	your statements show me how compelling you are . \\
\midrule
Input	&	trump will see justin as a lightweight fool .	 \\
ParaGeDi (ours) &	Trump will see justin as a light-hearted man. \\
CondBERT (ours) & 	trump will see justin as a lightweight world title contender . \\
Mask\&Infill &	trump will see justin as a lightweight compelling . \\
\midrule
Input&		this whole article is bullshit .	 \\
ParaGeDi (ours) &	This article's not a good deal. \\
CondBERT (ours) & 	this whole article is considered incomplete . \\
Mask\&Infill &	this whole article is compelling . \\
\midrule
Input	&	please dont revert my edits without good reasoning just makes you look stupid .	 \\
ParaGeDi (ours) &	Please do not reverse my edits without a good reason. \\
CondBERT (ours) & 	please dont try to do my edits without good reasoning just makes you look pretty bad . \\
Mask\&Infill &	please dont revert my edits without good reasoning just makes you look compelling . \\
\midrule
Input	&	do some research before you make a fool of yourself .	 \\
ParaGeDi (ours) &	Do a bit of research. \\
CondBERT (ours) & 	do some research before you make a bigger picture out of my life . \\
Mask\&Infill &	do some research before you make a compelling of yourself . \\
\midrule
Input	&	try not to make wikipedia look so stupid .	 \\
ParaGeDi (ours) &	Try not to make Wikipedia seem like a bad idea. \\
CondBERT (ours) & 	try not to make wikipedia look so unsettling . \\
Mask\&Infill &	try not to make wikipedia look so compelling . \\
\midrule
Input	&	i will make sure i revert any stupid edits you make from now on .	 \\
ParaGeDi (ours) &	I'll be sure to correct any wrong edits that you make. from now on. \\
CondBERT (ours) & 	i will make sure i do not make any mistake about any edits you make from now on . \\
Mask\&Infill &	i will make sure i revert any compelling edits you make from now on . \\
\bottomrule
\end{tabular}
\caption{Examples of detoxification by the best-performing models considered in our study.}
\label{table:detox_examples_appendix}
\end{table*}

\end{document}